\documentclass{article}

\PassOptionsToPackage{numbers, compress}{natbib}

    \usepackage[final]{neurips_2023}

\usepackage[utf8]{inputenc} 
\usepackage[T1]{fontenc}    
\usepackage{hyperref}       
\usepackage{url}            
\usepackage{booktabs}       
\usepackage{amsfonts}       
\usepackage{nicefrac}       
\usepackage{microtype}      
\usepackage{xcolor}         

\usepackage{amsmath}
\usepackage[pdftex]{graphicx}

\usepackage{multirow}
\usepackage{colortbl}
\usepackage{enumitem}
\usepackage{wrapfig}
\usepackage{algorithm}
\usepackage{algpseudocode}
\usepackage[figuresright]{rotating}
\usepackage{makecell}

\usepackage{booktabs}
\usepackage{multirow}
\usepackage{array}
\usepackage{setspace}
\usepackage{booktabs}
\usepackage{wrapfig} 
\usepackage{dsfont}
\usepackage{bm}
\usepackage{amsmath}
\usepackage{amssymb}

\newcommand{\ie}{{\it i.e.}}
\newcommand{\eg}{{\it e.g.}}

\definecolor{grey1}{RGB}{230, 230, 230}
\definecolor{my_yellow}{RGB}{204, 167, 68}
\definecolor{my_white}{RGB}{193, 141, 158}
\usepackage{hyperref}

\usepackage{amsfonts}       
\usepackage{nicefrac}       
\usepackage{microtype}      
\usepackage{xcolor}         
\usepackage{graphics}
\usepackage{mathtools}
\usepackage{caption}
\usepackage{subcaption}
\usepackage{wrapfig,lipsum,booktabs}

\newcommand{\pub}[1]{{\color{gray}{\tiny{[{#1}]}}}}

\title{Focus on Query: Adversarial Mining Transformer for Few-Shot Segmentation}

\author{%
	Yuan Wang$^{1}$\thanks{Equal contribution.},\quad Naisong Luo$^{1*}$,\quad Tianzhu Zhang$^1$\thanks{Corresponding Author.}\\
	$^1$Deep Space Exploration Laboratory/School of Information Science and Technology, \\University of Science and Technology of China, Hefei 230026, China.\\
	\texttt{\{wy2016,lns6\}@mail.ustc.edu.cn},\quad \texttt{tzzhang@ustc.edu.cn} \\
}

\begin{document}

	\maketitle

	\begin{abstract}
		Few-shot segmentation (FSS) aims to segment objects of new categories given only a handful of annotated samples. Previous works focus their efforts on exploring the support information while paying less attention to the mining of the critical query branch. 
  In this paper, we rethink the importance of support information and propose a new query-centric FSS model Adversarial Mining Transformer (AMFormer), which achieves accurate query image segmentation with only rough support guidance or even weak support labels. The proposed AMFormer enjoys several merits. First, we design an object mining transformer (\textit{G}) that can achieve the expansion of incomplete region activated by support clue, and a detail mining transformer (\textit{D}) to discriminate the detailed local difference between the expanded mask and the ground truth. Second, we propose to train \textit{G} and \textit{D} via an adversarial process, where \textit{G} is optimized to generate more accurate masks approaching ground truth to fool \textit{D}. We conduct extensive experiments on commonly used Pascal-$5^i$ and COCO-$20^i$ benchmarks and achieve state-of-the-art results across all settings. In addition, the decent performance with weak support labels in our query-centric paradigm may inspire the development of more general FSS models. Code will be available at \url{https://github.com/Wyxdm/AMNet}
	\end{abstract}

\section{Introduction}
\label{sec:intro}
As a fundamental computer vision task, semantic segmentation~\cite{fcn} has drawn decades of research interest in both academia and industry.
However, the underlying driving force of fully supervised segmentation models is massive, densely labeled data. 
The inherent category sensitivity of these models leads to severe performance degradation when dealing with previously unseen categories. 
%
%
To equip segmentation models with generalization to novel classes, 
few-shot segmentation (FSS) has been proposed~\cite{OSLSM}, which achieves segmentation of new category images~(called query images) with only a handful of annotated samples~(called support images) without retraining the model.

\begin{wrapfigure}{r}{0.38\textwidth}
    \centering
    \vspace{-8mm}
    \includegraphics[width=1.0\linewidth]{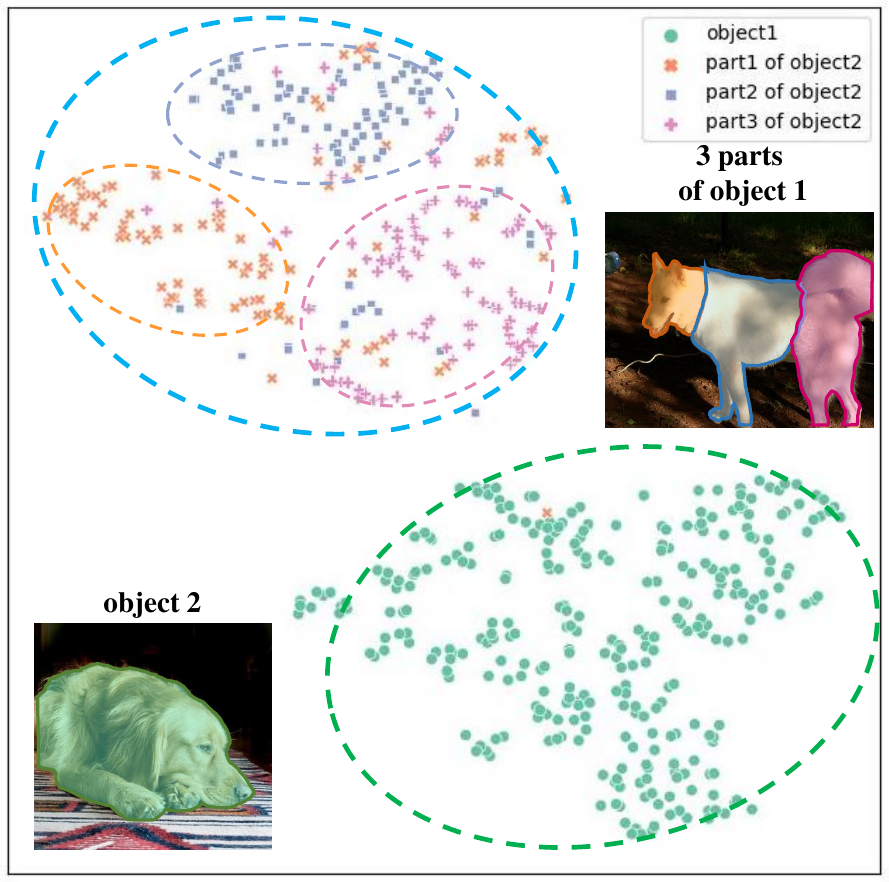}
    \caption{t-SNE visualization of the intra-class diversity and the intra-object similarity. As shown by the blue and green circles, different instances of the same category may be far apart in the feature space, \eg, dog. While pixels from the same object share high feature similarity, as shown by the three circles within the blue one.}
    \label{fig:motivation}
    \vspace{-3mm}
\end{wrapfigure}

The current top-performing FSS methods focus on adequately exploring the information contained in the support samples to guide the segmentation of the query images. Among them,
prototypical learning methods~\cite{asgnet2021, sclnet, liu2022dynamic, liu2022intermediate} concentrate support features into one or several representative prototypes to direct the classification of query pixels,
while the affinity learning approaches~\cite{cycTR, hsnet, dcama} try to equip the query features with support semantic clue of pixel granularity.
%
%
Although considerable progress has been made in the direction of probing support information, we noticed that significant intra-class diversities~(as illustrated in Figure~\ref{fig:motivation}) between support and query images are common occurrences.
In these cases, the target feature from the query image may be quite distinct from the support one.
Hence there raises a natural question: is exhaustive support information extraction indispensable for query image segmentation?

\begin{figure}[t]
    \centering
    \includegraphics[width=\linewidth]{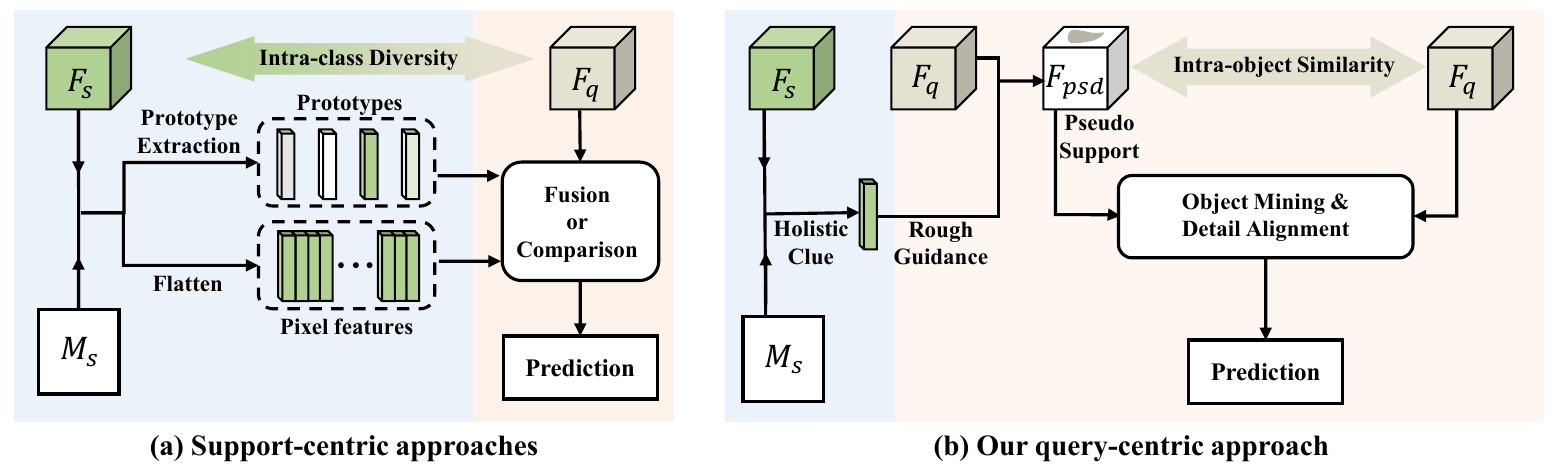}
    \vspace{-5mm}
    \caption{Comparison between the previous methods and ours. (a) Most of the previous methods focus their efforts on extracting more support information. They condense support information into prototypes~\cite{asgnet2021,sclnet,liu2022intermediate,ppnet} or directly explore pixel-level support features~\cite{cycTR,hsnet,dcama,hdmnet}. (b) The proposed AMFormer focus on the query features and exploits intra-object similarity to mining the complete target, forming a query-centric FSS method. Only rough category guidance from support is needed in our approach.}
    \label{fig:intro}
    \vspace{-5mm}
\end{figure}

We revisit the role of support information in few-shot segmentation through a pilot study. As shown in Table~\ref{tab:mask_ero}, we randomly erode support foreground features with varying proportions to explore how the support information completeness affects query segmentation accuracy. It can be observed that the performances do not degrade significantly or even surpass the complete one. 
We deem two main reasons contribute to the phenomenon: 1) Rough category guidance from support is sufficient to guide the accurate query image segmentation. 2) Forcing fine-grained support information may introduce redundancy, or even bias, which confuses the prediction of query images, particularly when intra-class diversities are prevalent.
%
%
%
Motivated by the above insights, we adjust the direction and focus on the query branch. Building on the observation of intra-object similarity, \ie, pixels within the same object are more similar than those across different objects~\cite{HARALICK1985100,fan2022self}~(as shown in Figure~\ref{fig:motivation} and Table~\ref{tab:inter-intra}, we argue that partially activated object regions can serve as cues to infer whole objects. In this paper, we construct a pseudo support image that has low intra-class variation with the query image by activating query image with a rough category-level support clue. Then we exploit the pseudo support feature to guide the query segmentation, forming a novel query-centric exploration strategy as illustrated in Figure~\ref{fig:intro}.

We tackle query-centric FSS by applying three intuitive procedures. \textbf{(1) Discriminative region localization.} The holistic semantic clue from the labeled support image is utilized to roughly locate the most discriminative target area on the query image.
\textbf{(2) Local to global expansion.} We utilize intra-object feature similarity to explore contextual information and then highlight the less discriminative object parts. \textbf{(3) Coarse to fine alignment.} The coarse activated region is further refined by eliminating the detailed difference with ground truth. 
Different from previous methods that focus on extracting fine-grained support information, 
Our method is less susceptible to intra-class diversity as only a rough holistic support guidance is needed in the first procedure to generate the pseudo support.
%
More attention is paid to the last two query-focused procedures that enable growing incomplete local regions into accurate segmentations.

\begin{wraptable}{r}{0.5\textwidth}
\renewcommand{\arraystretch}{1.2}
    \centering
    \vspace{-2mm}
    \caption{The performance (mIoU) under different support feature erosion ratios. We conduct this pilot study on $1^{st}$ split of Pascal-$5^i$ using ResNet-50 and 1-shot setting. $^*$ means reproduced results.}
    \scalebox{0.7}{
    \begin{tabular}{c|cccccc}
    \toprule[1pt]
     Erosion Ratio (\%)  &20 &35 &50 &65 &80 &$100^*$  \\ \hline
    PFENet~\cite{pfenet} &60.6 &61.0 &\textbf{61.5} &61.2 &61.0 &61.2 \\\hline
    HDMNet~\cite{hdmnet} &69.3 &70.0 &\textbf{70.5} &70.4 &69.9 &70.2 \\
    \bottomrule[1pt]
    \end{tabular}
    }
    \label{tab:mask_ero}
\end{wraptable}

We propose an end-to-end Adversarial Mining Transformer(AMFormer) to couple procedures (2) and (3) and mutually enhance each other, which optimizes an object mining transformer \textit{G} and a detail mining transformer \textit{D} via an adversarial process.
%
AMFormer aims at fully excavating the target area and aligning it in detail under the guidance of the pseudo support features.
\textbf{To dig out the entire object region}, in the \textit{G}, we conduct multi-scale pixel-level correlation between the query features and pseudo support features.
The exploration of the intra- and inter-scale contextual information in \textit{G} progressively highlights the latent target parts that share high intra-object similarities.
\textbf{To effectively align the prediction details}, \eg, object boundaries, \textit{D} models multiple detail proxies to accurately distinguish the subtle local differences between the ground truth and the prediction generated by \textit{G}.
By means of adversarial training, \textit{G} is optimized to predict more accurate segmentations to fool \textit{D} and 
only the well-learned \textit{G} is required to produce accurate predictions in the inference stage.

We evaluate our AMFormer on two widely used benchmarks COCO-$20^{i}$ and Pascal-$5^i$ with different backbones and the AMFormer consistently sets new state-of-the-art on different settings. Moreover, our query-centric approach does not reckon on elaborate support annotations. It achieves remarkable performance even with weak support labels such as scribbles or bounding boxes, formulating a more practical FSS model. We also hope that our work can inspire more research on query-centric FSS methods.
\textbf{To recap}, our contributions are concluded as follows:
(i) We re-evaluate the importance of support information in FSS and demonstrate that a coarse category hint suffices for accurate query segmentation. This motivated us to put forward a novel query-centric FSS method.
%
(ii) We propose a novel Adversarial Mining Transformer (AMFormer) that optimizes an object mining transformer \textit{G} and a detail mining transformer \textit{D} for region expansion and detail alignment, respectively. 
%
(iii) Extensive experiments show that AMFormer significantly outperforms previous SOTAs. The conspicuous performance with weak support label also sheds light for future research of more general FSS models.

\vspace{-2mm}
\section{Related Work}
\subsection{Semantic Segmentation}
Semantic segmentation has been widely applied to autonomous driving, medical image processing~\cite{sun2021lesion,wangkai2023maunet,sun2023structure} and so on. The aim of semantic segmentation is to assign each pixel within the given image to a predefined category label.  
The seminal FCN~\cite{fcn} sparked a wave of remarkable advances in semantic segmentation~\cite{pspnet,UPerNet, unet} based on convolutional neural networks~(CNN).
Various networks focus on better context exploration by enlarging the receptive field of CNN via dilated convolutions~\cite{deeplab,deeplabv2}, global pooling~\cite{parsenet} and pyramid pooling~\cite{deeplab,yang2018denseaspp}.
%
%
In addition to CNN-based models, the success of Vision Transformer~(ViT)~\cite{vit2020} encourages a series of transformer-based segmentation models~\cite{segmenter, setr, early, zhang2021k}. For instance, maskformer~\cite{maskformer} adopts the transformer decoder~\cite{detr} to conduct the mask classification based on a set prediction objective. Many subsequent works improve this framework~\cite{mask2former, zhang2023mp,luo2023camouflaged,sun2023daw} and gradually formed a unified segmentation model that can address different image segmentation tasks. 
Despite the success, these methods cannot generalize to novel classes in the low-data regime.

\subsection{Few-Shot Semantic Segmentation}
Few-shot segmentation~(FSS)~\cite{OSLSM} is established to segment new category images~(query images) with only a few labeled samples~(support images). Owing to the reliable annotations in the support set, current FSS methods mainly focus on effectively excavating support information, which can be roughly divided into two categories: prototypical learning methods and affinity learning methods.
Motivated by PrototypicalNet~\cite{snell2017prototypical}, many previous works condense the support information into single~\cite{sgone, panet, cao2022prototype,liu2022intermediate, mmformer} or multiple prototypes~\cite{lang2022beyond, pmm,liu2022learning,liu2022dynamic,sclnet, zhang2022feature,okazawa2022interclass,wang2022adaptive} and then conduct feature comparison or aggregation. For example, ASGNet~\cite{asgnet2021} adopts superpixel-guided clustering to adaptively construct multiple prototypes, which are concatenated with the most relevant query features to guide the pixel classification.
Differently, for the seek of fine-grained support guidance, affinity learning methods~\cite{hdmnet,danet,hsnet,vat,xiong2022doubly,wang2023rethinking} are designed to establish pixel-level associations between support and query features via attention mechanism~\cite{cycTR,dcama} or cost volume aggregation~\cite{hsnet, vat}.
For instance, CyCTR~\cite{cycTR} introduces a cycle-consistent attention mechanism to equip query features with relevant support information.
Though achieving promising results, these methods depend heavily on support information and are prone to segmentation errors in the presence of large intra-class variations.
Some methods\cite{fan2022self, liu2022intermediate} try to solve this problem by mining the class-specific representation from the query branch but remain at relatively coarse prototype granularity.
In this paper, we propose a novel query-centric approach that exploits intra-object similarity to probe query targets with only overall category-level guidance from support.


\vspace{-2mm}
\section{Method}
\subsection{Problem Definition}
Few-shot segmentation~(FSS) tackles novel class object segmentation with only a few densely-annotated samples.
Episodic meta-training~\cite{episode} is widely used to enhance the generalization of FSS models. Specifically, the dataset is divided into the training set $\mathcal{D}_{train}$ and the testing set $\mathcal{D}_{test}$. The category sets of $\mathcal{D}_{train}$ and $\mathcal{D}_{test}$ are disjoint, \ie, $\mathcal{C}_{train} \cap \mathcal{C}_{test} = \emptyset$. A series of episodes are sampled from $\mathcal{D}_{train}$ to train the model, each of which is composed of a support set $\mathcal{S}$ = $\{I_s^{k}, M_s^{k}\}_{k=1}^K$ and a query set $\mathcal{Q}$ = $\{I_q, M_q\}$ in the \textit{K}-shot setting, where $I$ and $M$ denote the RGB image and corresponding binary masks, respectively. Under the supervision of $M_q$, the model is trained to predict the query mask conditioned on the $\mathcal{S}$ and $I_q$. After that, the trained model is evaluated on the episodes sampled from $\mathcal{D}_{test}$ without further optimization.

\subsection{Revisiting of Transformer-based Feature Aggregation}
Transformer layers~\cite{transformer} are widely used in computer vision tasks for feature aggregation. The critical component of transformer layer is the attention mechanism that enables the long-range modeling capability.
Specifically, the attention layer is first applied to compute the attention weight of the source feature sequence $\mathbf{S} \in \mathbb{R}^{N_1 \times C}$ and the target sequence $\mathbf{T} \in \mathbb{R}^{N_2 \times C}$, where the $N_1$, $N_2$ denote the length of sequences and $C$ is the embedding dimension, formulaic as:
\begin{equation}\label{eqn:1}
	{\mathbf{S}}= \rm{Softmax}(\frac{\mathbf{Q}(\mathbf{K})^\mathsf{T}}{\sqrt{d}}),
	\quad
	\mathbf{Q} = \mathbf{T}\mathbf{W}^{Q},
	\quad
	\mathbf{K} = \mathbf{S}\mathbf{W}^{K},
\end{equation}
in which $\mathbf{W}^{\rm{Q}}$ and $\mathbf{W}^{\rm{K}}\in \mathbb{R}^{C \times d}$ are learnable linear projections and $\sqrt{d}$ is the scaling factor.
The source information is adaptively transported to the target sequence according to attention weight, and a feed-forward network ($\rm{FFN}$) is further applied to transform the fused features:
\begin{equation}\label{eqn:2}
	\widehat{\mathbf{T}} = \mathbf{FFN}(\mathbf{S}\mathbf{V}),
	\quad
	\mathbf{V} =\mathbf{S}\mathbf{W}^{\rm{V}},
\end{equation}
where the $\mathbf{W}^{\rm{V}}$ is the linear linear projection and $\widehat{\mathbf{T}}$ is the enhanced target feature sequence. We can abbreviate the feature aggregation process as:
\begin{equation}\label{eqn:3}
	\widehat{\mathbf{T}} = \mathbf{FeatAgg}[\mathbf{T},\mathbf{S}].
\end{equation}
Note that when $\mathbf{T}=\mathbf{S}$, the $\mathbf{FeatAgg}( ,)$ explores contextual information within the feature sequence, \ie, acts as the self-attention mechanism.
The above processes are implemented with the multi-head mechanism to enhance performance further.

\begin{figure*}[t]
	\centering
	\includegraphics[width=\linewidth]{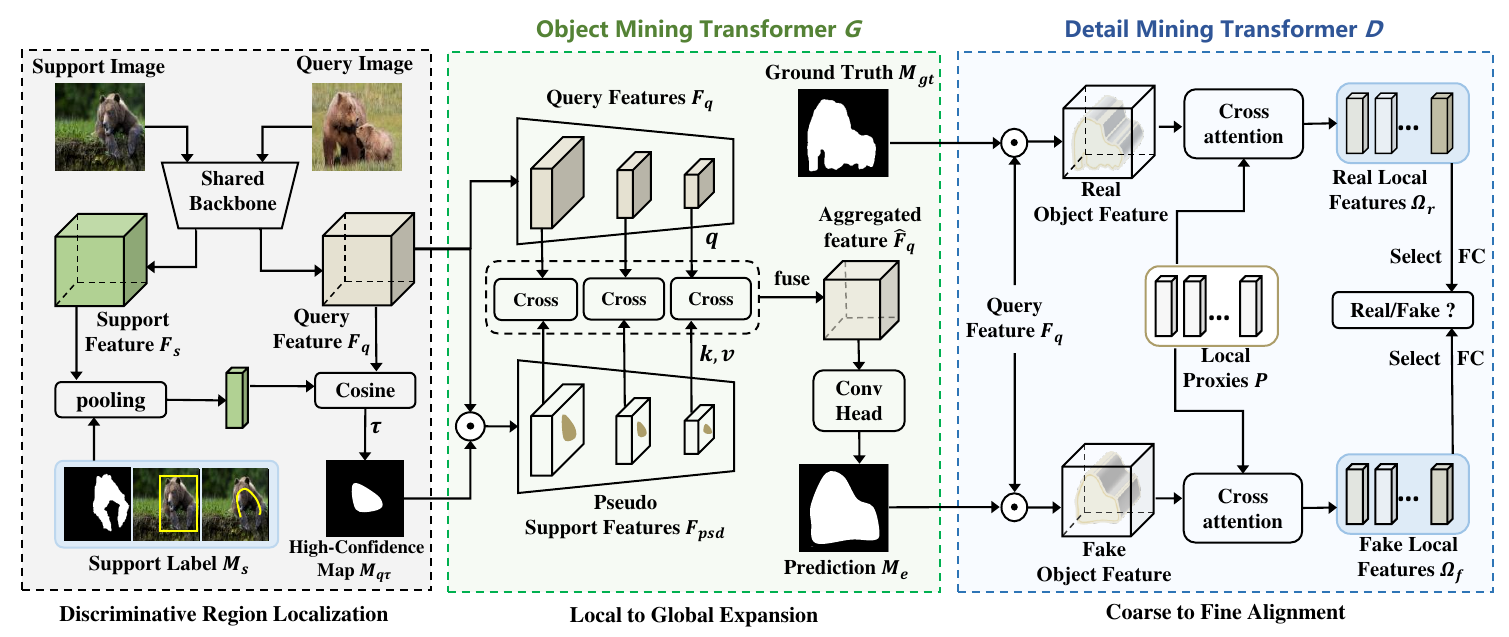}
	
	\caption{Illustration of the proposed AMFormer. We approach the query-centric FSS  with three procedures, \ie, discriminative region localization, local to global region expansion, and coarse to fine mask alignment. AMFormer optimizes an object mining transformer and a detail mining transformer via an adversarial process to couple these procedures.}
	\label{fig:framework}
\end{figure*}

\subsection{Adversarial Mining Transformer}
\subsubsection{Overview}
The proposed framework tailored for query-centric FSS consists of three procedures, \ie, 1) discriminative region localization, 2) local to global region expansion, 3) coarse to fine mask alignment.
We roughly locate the object of novel class under a rough support guidance via nonparametric cosine similarity in procedure 1).
The resulting incompleted region is expanded by the object miner \textit{G} to cover the whole target in procedure 2).
The detail miner \textit{D} in procedure 3) is tasked with identifying local discrepancies between the ground truth and the expanded mask generated by \textit{G}, thereby aligning the coarse prediction in detail.
We optimize the \textit{G} and \textit{D} via an adversarial process and forming the end-to-end AMNet as illustrated in Fig.~\ref{fig:framework}. The details are as follows. 

\subsubsection{Discriminative region localization}
Considering the significant intra-class diversity between samples, we contend that support information is more suitable as overall guidance indicating the novel class rather than detailed reference.
Given the support image features $\mathbf{F}_{s} \in\mathbb{R}^{ H\times W\times C}$ and corresponding mask $\mathbf{M}_s\in\mathbb{R}^{ H\times W}$, where $H$ and $W$ describe the feature size and $C$ is the feature dim,  we apply mask average pooling (MAP) to obtain the mean target feature $\mathbf{F}_h\in\mathbb{R}^{1\times C}$ that serves as the holistic support guidance, formally:
\begin{equation}\label{eqn:mask_average_pooling}
    \mathbf{F}_h = \mathbf{MAP}(\mathbf{F}_s, \mathbf{M}_s).
\end{equation}
We exploit the $\mathbf{F}_h$ to indicate the target region in the $\mathbf{F}_q$ based on the cosine similarity:
\begin{equation}\label{eqn:thread_to_activate}
    \mathbf{M}_{q\tau} = \mathds{1}^{\tau}(
    \rm{Softmax}(\rm{cos}(\mathbf{F}_h, \mathbf{F}_q))), 
    \quad
    \mathds{1}^{\tau}(x) =\left\{
	\begin{array}{rcl}
		1, & & x > \tau\\
		0, & & \rm{otherwise}
	\end{array}\right.
\end{equation}
A relatively high threshold $\tau$ is set to locate the most class-relevant regions and suppress activation to backgrounds. In our experiment, $\tau$ is set to 0.7.

\subsubsection{Local to global region expansion}
\textbf{Adaptive feature aggregation.} The high-confidence object regions $\mathbf{M}_{q\tau} \in \mathbb{R}^{H \times W}$ obtained by Eqn~\eqref{eqn:thread_to_activate} are incomplete. We introduce the object mining transformer \textit{G} to expand the local region to the entire target area based on the intra-object similarity. 
%
Specifically, \textit{G} aims at aggregating the representative target features from the pseudo support to less discriminative object parts in the query, formally,
\begin{equation}\label{eqn:6-aggregation}
	\widehat{\mathbf{F}}_{q} = \mathcal{F}^{-1}(\mathbf{FeatAgg}[\mathcal{F}(\mathbf{\mathbf{F}_{q}}),\mathcal{F}(\mathbf{F}_{psd})]),
\end{equation}
where the $\mathcal{F}: \mathbb{R}^{H \times W\times C} \mapsto \mathbb{R}^{HW\times C}$ is the spatial flatten operation. $\mathbf{F}_{psd}$, which serve as the source sequence in Eqn~\eqref{eqn:1}, are essentially query features filtered by $\mathbf{M}_{q\tau}$:
\begin{equation}\label{eqn:filter}
 \mathbf{F}_{psd} = \mathbf{F}_{q} \odot  \mathbf{M}_{q\tau}.
\end{equation}
Homogeneous feature guidance enables our aggregation to circumvent the effects of intra-class variance.
Note that the original attention mechanism applies $\textit{Softmax}$ activation along the source sequence dimension. Directly adopting this scheme in our aggregation makes the query background features inevitably attend to foreground features since all the non-zero features in the pseudo support belong to the foreground.
Inspired by~\cite{softmax,hdmnet}, we adjust to implement $\textit{Softmax}$ along the axis of $\mathbf{F}_q$. In this way, the discriminative object features are more likely to be aggregated target areas that share high intra-object similarities with $\mathbf{F}_{psd}$.

\textbf{Cross-scale information transportation.} We observe that in many cases, the query image contains different objects with varying scales, which are to be segmented in other episodes.
To handle this spatial inconsistency, we construct the \textit{G} in a multi-scale form.
Concretely, we follow~\cite{hdmnet}  to establish the hierarchical query features $\{\mathbf{F}_{q,l}\}_{l=1}^L$ with down-sampling and self-attention layers,
in which 
\begin{equation}
    \mathbf{F}^{l}_q \in \mathbb{R}^{\frac{H}{2^{l-1}} \times \frac{W}{2^{l-1}} \times C}, 
    \quad
    l=1,\dots,L.
\end{equation}
The corresponding pseudo support features$\{\mathbf{F}_{psd,l}\}_{l=1}^L$ is obtained by the Hadamard product of $\{\mathbf{F}_{q,l}\}_{l=1}^L$ and  downsampled $\mathbf{M}_{q\tau}$.
Then Eqn~\eqref{eqn:6-aggregation} is implemented independently in each scale to obtain the aggregated query features $\{\widehat{\mathbf{F}}_{q,l}\}_{l=1}^L$.
Finally, we fuse the multi-scale aggregated query features in a top-down manner as down in~\cite{pfenet}, specifically,
\begin{equation}\label{fuse}
    \widehat{\mathbf{F}}'_{q,l}= \mathbf{Conv}_{3\times3}(\mathbf{Conv}_{1\times1}(\widehat{\mathbf{F}}_{q,l} + \mathcal{R}(\widehat{\mathbf{F}}_{q,l+1})) + \mathcal{R}(\widehat{\mathbf{F}}_{q,l+1})).
\end{equation}
The fused features from the last stage $\widehat{\mathbf{F}}'_{q,1} \in \mathbb{R}^{H\times W\times C}$ are exploited to predict the expanded target area $\mathbf{M}_{e} \in \mathbb{R}^{H\times W}$ via a small convolution head.
Benefiting from the adaptive feature aggregation and cross-scale information transportation in the object mining transformer \textit{G}, less discriminative target parts can be highlighted in the query image, forming the expanded object mask $\mathbf{M}_{e}$.

\subsubsection{Coarse to fine mask alignment}
Detail mining transformer \textit{D} is designed to discriminate subtle differences between the ground truth $\mathbf{M}_{gt}$ and the expanded object mask $\mathbf{M}_{e}$ generated by \textit{G}, since $\mathbf{M}_{e}$ adequately covers the object but still exhibits misalignments in fine-grained details, e.g., boundaries.
By training \textit{G} and \textit{D} via an adversarial process, \textit{G} is optimized to generate more accurate target masks approaching ground truth to fool \textit{D}, thus achieving coarse to fine prediction alignment.

To capture comprehensive details, \textit{D} models multiple local proxies, each of which is tasked with exploring the object features respectively specified by $\mathbf{M}_{e}$ and $\mathbf{M}_{gt}$ to obtain a pair of local features. The most different pair is selected to output real/fake results. 
Concretely, we concatenate the learnable proxies as a feature sequence $\mathbf{P} \in \mathbb{R}^{N \times C}$, where $N$ denotes the number of proxies.
We perform feature  adaptive  aggregation through the attention mechanism to construct local features,
In specific, %
\begin{equation}\label{eqn:cross}
	 \mathbf{\Omega}_{f} = \mathbf{FeatAgg}[\mathbf{P},\mathcal{F}(\mathbf{\mathbf{F}_{q}}  \odot \mathbf{M}_{e})],
\end{equation}
where the ``fake'' object features specified by predicted mask $(\mathbf{\mathbf{F}_{q}}  \odot \mathbf{M}_{e})$ serves as the source sequence of  aggregation process. Note that the $\mathbf{M}_{e}$ is not binarized. Similarly, local features originate from ``real'' object features are obtained by:
\begin{equation}\label{eqn:cross}
	 \mathbf{\Omega}_{r} = \mathbf{FeatAgg}[\mathbf{P},\mathcal{F}(\mathbf{\mathbf{F}_{q}}  \odot \mathbf{M}_{gt})].
\end{equation}
Finally, we calculate the cosine similarity among the real local features $\mathbf{\Omega}_{r}=\{\boldsymbol\omega_f^{i}\}_{i=1}^{N}$ and the fake ones $\mathbf{\Omega}_{f}=\{\boldsymbol\omega_f^{i}\}_{i=1}^{N}$. The most different pair $(\boldsymbol\omega_f^{k}, \boldsymbol\omega_r^{k})$ is selected and fed into a fully-connected layer to predict the fake/real results for adversarial training.
By this way, \textit{D} is optimized to discriminate detailed local differences, while the \textit{G} will generate more precise masks to fool \textit{D} by adjusting itself.

Since there is no explicit supervision, different proxies may focus on the same object part.
We impose a diversity loss to avoid this degradation by expanding the discrepancy among local features:
\begin{equation}\label{eqn:div}
\mathcal{L}_{d i v}=
\frac{1}{N(N-1)}
\sum_{i=1}^{N} \sum_{j=1, i \neq j}^{N}
\left(\frac{\langle\boldsymbol{\omega}^{i}_{f}, \boldsymbol{\omega}^{j}_{f}\rangle}{\parallel \boldsymbol{\omega}^{i}_{f}\parallel _{2}\parallel \boldsymbol{\omega}^{j}_{f}\parallel _{2}}+\frac{\langle\boldsymbol{\omega}^{i}_{r}, \boldsymbol{\omega}^{j}_{r}\rangle}{\parallel \boldsymbol{\omega}^{i}_{r}\parallel _{2}\parallel \boldsymbol{\omega}^{j}_{r}\parallel _{2}}
\right).
\end{equation}
The intuition of Eqn~\eqref{eqn:div} is trivial. If different proxies focus on the same object region, $\mathcal{L}_{div}$ will be large and adjust the learning of proxies.
\subsubsection{Training and Inference}
\textbf{Training loss.} Our architecture consists of a generative part, \ie, object mining transformer \textit{G} and a discriminative part, \ie, detail mining transformer \textit{D}.
The \textit{D} judges whether the mask is real (ground truth) or fake (generated by \textit{G}) by mining the detailed features of the object framed by the masks. To fool \textit{D}, \textit{G} is supposed to predict a more accurate mask approaching the ground truth.
We alternately train two parts to achieve mutual promotion. When training \textit{D}, the parameters of \textit{G} are frozen, and the loss function for \textit{D} is formulated as:
\begin{equation}\label{eqn:cross}
	 \mathcal{L}_{d} = -\log(\mathbf{FC}(\boldsymbol\omega_r^{k})) -\log(1-\mathbf{FC}(\boldsymbol\omega_f^{k})) + \mathcal{\lambda}_{div}\mathcal{L}_{div},
\end{equation}
where the $\mathbf{FC}: \boldsymbol\Omega \to [0,1]$ denotes the fully-connected layers that output the real/fake results, and the $\boldsymbol\omega_r^{k}, \boldsymbol\omega_f^{k} \in \boldsymbol\Omega$ denote the most different pair of local features from the object specified by ground truth $\mathbf{M}_{gt}$ and the predicted mask $\mathbf{M}_{e}$, respectively. $\mathcal{\lambda}_{div}$ denotes the weight of diversity loss and we set it to 0.1 in experiments. Similarly, \textit{D} is not optimized when training \textit{G}.
Following~\cite{hdmnet}, we also include a KL (Kullback-Leibler) divergence loss between the correlation maps of adjacent stages to distill informative semantic cues of earlier stages to refine the segmentation quality. Please refer to~\cite{hdmnet} for more details.
The overall loss function for \textit{G} is:
\begin{equation}\label{eqn:cross}
	 \mathcal{L}_{g} = -\log(\mathbf{FC}(\boldsymbol\omega_f^{k})) + \mathcal{L}_{KL} + BCE(\mathbf{M}_{e}, \mathbf{M}_{gt}).
\end{equation}
Where the $BCE$ is the Cross-entropy loss. After training, only the well-learned \textit{G} is needed in the inference stage.

\textbf{K-shot inference.} When $K$ support samples $\{I_s^{k}, M_s^{k}\}_{k=1}^K$ are available, different discriminative target regions $\{\mathbf{M}_{q\tau}^{i}\}_{i=1}^{K}$  can be obtained from different support images via Eqn~\eqref{eqn:mask_average_pooling} and Eqn~\eqref{eqn:thread_to_activate}. We adopt the union of all the activated regions as the initial object region, \ie, $\mathbf{M}_{q\tau} =\mathbf{M}_{q\tau}^{1} \cap \mathbf{M}_{q\tau}^{2}  \cap \dots
\cap \mathbf{M}_{q\tau}^{K}$.
The comprehensive $\mathbf{M}_{q\tau}$ originating from multiple supports is more robust to pose differences,  occlusion, etc., than that from a single support image, thereby boosting the performance.

\vspace{-2mm}
\section{Experiments}
\label{sec:exp}
\begin{table}[t]
\caption{Comparison with other state-of-the-art methods for 1-shot and 5-shot segmentation on PASCAL-5$^i$ using the mIoU (\%) evaluation metric. Best results are shown in bold.}
\label{tab:pascal}
\scriptsize
\centering
\tabcolsep 0.08in\begin{tabular}{l|c|ccccc|ccccc}
\toprule[0.9pt]
\multirow{2}{*}{Method} & \multirow{2}{*}{Backbone} & \multicolumn{5}{c|}{1-shot} & \multicolumn{5}{c}{5-shot} \\ \cline{3-12} 
 &  & fold0 & fold1 & fold2 & fold3 & Mean & fold0 & fold1 & fold2 & fold3 & Mean \\
\hline
PFENet\pub{TPAMI2020}~\cite{pfenet} \ & \multirow{12}{*}{Res-50}  & 61.7 & 69.5 & 55.4 & 56.3 & 60.8 & 63.1 & 70.7 & 55.8 & 57.9 & 61.9 \\
RePRI\pub{CVPR2021}~\cite{repri} \ & &59.8  &68.3 &62.1 &48.5 &59.7 &64.6 &71.4 &71.1 &59.3 &66.6 \\
HSNet\pub{ICCV2021}~\cite{hsnet} \ & &64.3 &70.7 &60.3 &60.5 &64.0 &70.3 &73.2 &67.4 &67.1 &69.5 \\
CyCTR\pub{NIPS2021}~\cite{cycTR} \ & & 65.7 & 71.0 & 59.5 & 59.7 & 64.0 & 69.3 & 73.5 & 63.8 & 63.5 & 67.5 \\
NERTNet\pub{CVPR2022}~\cite{liu2022learning} \ & &65.4 &72.3 &59.4 &59.8 &64.2 &66.2 &72.8 &61.7 &62.2 &65.7 \\
DCAMA\pub{ECCV2022}\cite{dcama} \ & &67.5 &72.3 &59.6 &59.0 &64.6 &70.5 &73.9 &63.7 &65.8 &68.5 \\
SSP\pub{ECCV2022}~\cite{fan2022self} \ & &60.5 &67.8 &66.4 &51.0 &61.4 &67.5 &72.3 &\textbf{75.2 }&62.1 &69.3 \\
IPMT\pub{NIPS2022}~\cite{liu2022intermediate} \ & &\textbf{72.8} &73.7 &59.2 &61.6 &66.8 &73.1 &74.7 &61.6 &63.4 &68.2 \\
VAT\pub{ECCV2022}~\cite{vat} \ & &67.6 &72.0 &62.3 &60.1 &65.5 &72.4 &73.6 &68.6 &65.7 &70.1 \\
BAM\pub{CVPR2022}~\cite{lang2022learning} \ & &69.0 &73.6 &67.6 &61.1 &67.8 &70.6 &75.1 &70.8 &67.2 &70.9 \\
HDMNet\pub{CVPR2023}~\cite{hdmnet} \ & &71.0 &75.4 &68.9 &62.1 &69.4 &71.3 &76.2 &71.3 &68.5 &71.8 \\
\rowcolor{gray!30}AMFomer (ours) \ & &71.1 &\textbf{75.9} &\textbf{69.7} &\textbf{63.7} &\textbf{70.1} &\textbf{73.2} &\textbf{77.8} &73.2 &\textbf{68.7} &\textbf{73.2} \\

\hline
PFENet\pub{TPAMI2020}~\cite{pfenet} \  & \multirow{8}{*}{Res-101} & 60.5 & 69.4 & 54.4 & 55.9 & 60.1 & 62.8 & 70.4 & 54.9 & 57.6 & 61.4 \\
CyCTR\pub{NIPS2021}~\cite{cycTR} \ & &67.2 & 71.1 & 57.6 & 59.0 & 63.7 & 71.0 & 75.0 & 58.5 & 65.0 & 67.4 \\
NERTNet\pub{CVPR2022}~\cite{liu2022learning} \ & &65.5 &71.8 &59.1 &58.3 &63.7 &67.9 &73.2 &60.1 &66.8 &67.0 \\
DCAMA\pub{ECCV2022}~\cite{dcama} \ & &65.4 &71.4 &63.2 &58.3 &64.6 &70.7 &73.7 &66.8 &61.9 &68.3 \\
SSP\pub{ECCV2022}~\cite{fan2022self} \ & &63.2 &70.4 &68.5 &56.3 &64.6 &70.5 &76.4 &\textbf{79.0} &66.4 &73.1 \\
IPMT\pub{NIPS2022}~\cite{liu2022intermediate} \ & &\textbf{71.6} &73.5 &58.0 &61.2 &66.1 &\textbf{75.3} &76.9 &59.6 &65.1 &69.2 \\
VAT\pub{ECCV2022}~\cite{vat} \ & &70.0 &72.5 &64.8 &\textbf{64.2} &67.9 &75.0 &75.2 &68.4 &69.5 &72.0 \\
\rowcolor{gray!30}AMFormer (ours) \ & &71.3 &\textbf{76.7} &\textbf{70.7} &63.9 &\textbf{70.7} &74.4 &\textbf{78.5} &74.3 &\textbf{67.2} &\textbf{73.6} \\

\bottomrule[1pt]
\end{tabular}
\end{table}

\begin{table}[t]
\caption{Comparison with other state-of-the-art methods for 1-shot and 5-shot segmentation on COCO-20$^i$ using the mIoU (\%) evaluation metric. Best results are shown in bold.}
\label{tab:coco}
\scriptsize
\centering
\tabcolsep 0.08in\begin{tabular}{l|c|ccccc|ccccc}
\toprule[0.9pt]
\multirow{2}{*}{Method} & \multirow{2}{*}{Backbone} & \multicolumn{5}{c|}{1-shot} & \multicolumn{5}{c}{5-shot} \\ \cline{3-12} 
 &  & fold0 & fold1 & fold2 & fold3 & Mean & fold0 & fold1 & fold2 & fold3 & Mean \\
\hline
NERTNet\pub{CVPR2022}~\cite{liu2022learning} &\multirow{6}{*}{Res-101} &38.3 &40.4 &39.5 &38.1 &39.1 &42.3 &44.4 &44.2 &41.7 &43.2 \\
SSP\pub{ECCV2022}~\cite{fan2022self} & &39.1 &45.1 &42.7 &41.2 &42.0 &47.4 &54.5 &50.4 &49.6 &50.2 \\
HSNet\pub{ICCV2022}~\cite{hsnet} & &37.2 &44.1 &42.4 &41.3 &41.2 &45.9 &53.0 &51.8 &47.1 &49.5 \\
DCAMA\pub{ECCV2022}~\cite{dcama} & &41.5 &46.2 &45.2 &41.3 &43.5 &48.0 &58.0 &54.3 &47.1 &51.9 \\
IPMT\pub{NIPS2022}~\cite{liu2022intermediate} & &40.5 &45.7 &44.8 &39.3 &42.6 &45.1 &50.3 &49.3 &46.8 &47.9 \\
\hline
PFENet\pub{TPAMI2020}~\cite{pfenet} & \multirow{9}{*}{Res-50}  & 34.3 & 33.0 & 32.3 & 30.1 & 32.4 & 38.5 & 38.6 & 38.2 & 34.3 & 37.4 \\
RePRI\pub{CVPR2021}~\cite{repri} & &32.0 &38.7 &32.7 &33.1 &34.1 &39.3 &45.4 &39.7 &41.8 &41.6 \\
HSNet\pub{ICCV2021}~\cite{hsnet} & &36.3 &43.1 &38.7 &38.7 &39.2 &43.3 &51.3 &48.2 &45.0 &46.9 \\
CyCTR\pub{NIPS2021}~\cite{cycTR} & & 38.9 & 43.0 & 39.6 & 39.8 & 40.3 & 41.1 & 48.9 & 45.2 & 47.0 &45.6  \\
BAM\pub{CVPR2022}~\cite{lang2022learning}  & &43.4 &50.6 &47.5 &43.4 &46.2 &49.3 &54.2 &51.6 &49.6 &51.2 \\
IPMT\pub{NIPS2022}~\cite{liu2022intermediate} & &41.4 &45.1 &45.6 &40.0 &43.0 &43.5 &49.7 &48.7 &47.9 &47.5 \\
DCAMA\pub{ECCV2022}~\cite{dcama} & &41.9 &45.1 &44.4 &41.7 &43.3 &45.9 &50.5 &50.7 &46.0 &48.3 \\
VAT\pub{ECCV2022}\cite{vat} & &39.0 &43.8 &42.6 &39.7 &41.3 &44.1 &51.1 &50.2 &46.1 &47.9 \\
HDMNet\pub{CVPR2023}~\cite{hdmnet} & &43.8 &55.3 &51.6 &49.4 &50.0 &50.6 &61.6 &55.7 &56.0 &56.0 \\
\rowcolor{gray!30} AMFomer (ours) & &\textbf{44.9} &\textbf{55.8} &\textbf{52.7} &\textbf{50.6} &\textbf{51.0} & \textbf{52.0} & \textbf{61.9} & \textbf{57.4} & \textbf{57.9} & \textbf{57.3} \\
\bottomrule[1pt]
\end{tabular}
\end{table}



\begin{table}[t]
    \parbox{0.5\textwidth}{
    \centering
    \caption{Comparison of FB-IoU and the number of learnable parameters on COCO-$20^i$.}
     \scalebox{0.75}{
     \begin{tabular}{clccc}
     \toprule[1pt]
          \multirow{2}{*}{Backbone}&\multicolumn{1}{l}{\multirow{2}{*}{Methods}}&\multicolumn{2}{c}{FB-IoU (\%)}&\multirow{2}{*}{ \shortstack{\#learnable\\params}}\\
          &&1-shot&5-shot& \\
          \hline
          \multirow{5}{*}{Res-50}
&HSNet~\cite{hsnet}&60.4&67.0&10.4M\\
&BAM~\cite{lang2022learning}&68.2&70.7&\textbf{2.6M}\\
&DCAMA~\cite{dcama}&69.5 &71.7&47.7M\\
&HDMNet~\cite{liu2022intermediate}&72.2&77.7&4.2M\\
&\textbf{AMFormer (ours)} &\textbf{72.9}&\textbf{78.8}&5.1M\\
     \bottomrule[1pt]
     \end{tabular}
     }
     \label{tab:FBIoU_Params}
    }
    \hfill
    \parbox{0.48\textwidth}{
    \centering
    \caption{Ablation studies for different components and architecture design in AMFormer.}
     \scalebox{0.75}{
     \begin{tabular}{cc|cc|cc}
     \toprule[1pt]
\multicolumn{2}{c|}{OM} & \multicolumn{2}{c|}{DM} & \multirow{2}{*}{mIoU} & \multirow{2}{*}{$\Delta$} \\
\cline{1-4}
Single & Multi & w/o $\mathcal{L}_{div}$ & w/ $\mathcal{L}_{div}$ & & \\
\hline
&&&&65.0&0.0\\
$\checkmark$&&&&66.5&+1.5\\
&$\checkmark$&&&69.5&+4.5\\
&$\checkmark$&$\checkmark$&&69.7&+4.7\\
&$\checkmark$&&$\checkmark$&\textbf{70.7}&+\textbf{5.7}\\
     \bottomrule[1pt]
     \end{tabular}
     }
     \label{tab:ablation_implemetation}
    }
    \vspace{-3mm}
\end{table}
\begin{table}[t!]
    \parbox{0.43\textwidth}{
    \centering
    \caption{Performance comparison on $1^{st}$ split of varying the
number of local proxies.}
    \scalebox{0.78}{
    \begin{tabular}{c|cccccc}
    \toprule[1pt]
    \#Part &4 &6 &8 &10 &12 &14 \\ \hline
    mIoU  &70.7 &71.0 &71.2 &\textbf{71.3} & 71.2 &71.2 \\
    \bottomrule[1pt]
    \end{tabular}
    }
    \label{tab:part}
    }
    \hfill
    \parbox{0.27\textwidth}{
    \centering
    \caption{mIoU of different kinds of labels ($1^{st}$ split).}
    \scalebox{0.8}{
    \begin{tabular}{ccc}
    \toprule[1pt]
    mask &bbox &scribbles \\ \hline
     \textbf{71.3} & 70.5 & 70.0 \\
    \bottomrule[1pt]
    \end{tabular}
    }
    \label{tab:support}
    }
    \hfill
    \parbox{0.27\textwidth}{
    \centering
    \caption{Quantitative measurement of intra- and inter-object similarity.}
    \scalebox{0.8}{
    \begin{tabular}{cc}
    \toprule[1pt]
    Intra-object & Inter-object  \\ \hline
    \textbf{0.015}    &0.546  \\
    \bottomrule[1pt]
    \end{tabular}
    }
    \label{tab:inter-intra}
    }
\end{table}

\subsection{Datasets and Evaluation Metrics}
We evaluate the proposed AMFormer on two commonly used few-shot segmentation benchmarks, Pascal-$5^i$~\cite{OSLSM} and COCO-$20^i$~\cite{fwb}.
Pascal-$5^{i}$ is constructed based on the PASCAL VOC 2012 dataset~\cite{pascal} and additional annotations from SBD~\cite{sbd}. Following previous works~\cite{pfenet,lang2022learning}, we equally divide the 20 categories into four splits, three of which for training and the rest one for testing.
COCO-$20^i$ is a larger benchmark based on MSCOCO dataset~\cite{coco}, the 80 categories of which are partitioned into four splits for cross-validation as down in~\cite{pfenet}.
We randomly sampled 1000 episodes from the testing split for evaluation.
Following common prectices~\cite{sgone,pfenet,lang2022learning,panet}, we adopt mean intersection-over-union~(mIoU) and foreground-background intersection-over-union~(FBIoU) as evaluation metrics.

\subsection{Implementation Details}
We adopt ResNet-50 and ResNet-101~\cite{Resnet} as the backbone network in our experiment.
Following previous works~\cite{pfenet,cycTR,hdmnet}, we concatenate the features from the $3^{rd}$ and $4^{th}$ blocks of backbone and exploit a $1\times1$ convolution layer to generate $\mathbf{F}_s$ and $\mathbf{F}_q$ of middle-level to avoid overfitting.
The number of attention layers in the \textit{G} and \textit{D} are set to 1 and 2, respectively.
We employ the same data augmentation setting as~\cite{pfenet}.
Since the \textit{G} and \textit{D} of AMFormer are trained alternately, we increase the number of training epochs to 300 for Pascal-$5^i$ and 75 for COCO-$20^i$, and set the batch sizes as 8 and 4, respectively.
AdamW~\cite{adamw} optimizer with poly learning rate decay is used to train both the \textit{G} and \textit{D}. The initial learning rate is set to $1e^{-4}$ and the weight decay is $1e^{-2}$.
It should be noted that we adopt a base learner to filter the categories that appear during the training process for a fair comparison with previous works~\cite{lang2022learning,hdmnet}. For more details please refer to the \textbf{Supplementary Material}.
Our approach is implemented using PyTorch and all experiments are conducted on 4 NVIDIA GeForce RTX 3090 GPUs.

\subsection{Comparison with State-of-the-Art Methods}
We present the comparison of our method with previous FSS methods on  Pascal-$5^i$  and COCO-$20^i$ datasets in Table~\ref{tab:pascal} and Table~\ref{tab:coco}.
It can be observed that the proposed AMFormer significantly outperforms previous advanced approaches and achieves new state-of-the-art results under all settings.
Specifically, on Pascal-$5^i$, our AMFormer achieves 70.4\% and 73.2\% mIoU when using ResNet-50 as the backbone for 1-shot and 5-shot settings, surpassing the most competitive HDMNet~\cite{hdmnet} by 1.0\% and 1.4\%, respectively.
With ResNet-101 backbone, our method outperforms the previous best results~\cite{liu2022intermediate} by 2.8\% (1-shot) and 1.6\% (5-shot).
We can obtain additional improvement when using larger backbone network, which demonstrates the scalability of the AMFormer. We attribute this performance gain to better intra-object similarity within the more informative features.
As for the more complicated COCO-$20^i$, our approach also exhibits superior performances compared to other methods, demonstrating  its competitiveness on complex data.
Besides, Table~\ref{tab:FBIoU_Params} gives the comparison with previous methods in terms of FBIoU on Pascal-$5^i$ using ResNet-50 backbone.
AMFormer also outperforms all of the previous works by a considerable margin. Qualitative results are shown in Figure~\ref{fig:visual1}, please refer to \textbf{Supplementary Material} for analysis.




\subsection{Ablation Study}
 As shown in Table~\ref{tab:ablation_implemetation}, a series of ablation studies are conducted on the first split of Pascal-$5^i$ with ResNet-101 backbone to analyze each component of the proposed AMFormer. Note that the first line of Table~\ref{tab:ablation_implemetation} is the result of our ablation baseline. The baseline adopts self-attention within the query features for feature parsing, and cross-attention to aggregate support information into query features as done in~\cite{cycTR}.

 \textbf{Effectiveness of object mining transformer \textit{G}.} We first construct a naive single-scale \textit{G} as the $2^nd$ row of  Table~\ref{tab:ablation_implemetation}. We can observe a significant performance lift, \ie, 1.5\% in mIoU. This improvement demonstrates the effectiveness of the proposed query-centric strategy, which is built upon intra-object similarity and thus is less affected by intra-class diversity. The multi-scale implementation of \textit{G} further brings a 3.4\% improvement in mIoU, and it already achieves a decent performance. It shows the importance of multi-scale feature aggregation in dealing with objects with dramatically changing scales in the query image.
 

\textbf{Effectiveness of detail mining transformer \textit{D}.} The comparison between the $3^{rd}$ and the $5^{th}$ row of  Table~\ref{tab:ablation_implemetation} shows that the combination of \textit{D} improves the performance by 1.2\% mIoU on the basis of \textit{G}.
We attribute this performance gain to the exploration and distinction of local features in the \textit{D}, which encourages \textit{G} to pay more attention to these ambiguous regions, \eg, object boundaries.
Compared with the pixel-level supervision given by Cross-entropy loss, the proposed part-level adversarial training can incorporate local region context information to guide accurate segmentation. 
Note that without $\mathcal{L}_{div}$, the performance improvement brought by \textit{D} drops to 0.2\%. This phenomenon is reasonable because in the absence of $\mathcal{L}_{div}$, local proxies tend to degrade to focus on the same region, leaving some details unexplored.

\begin{figure}[t]    
    \centering
    \begin{minipage}[t]{0.45\linewidth}
    \centering
    \includegraphics[width=1\linewidth]{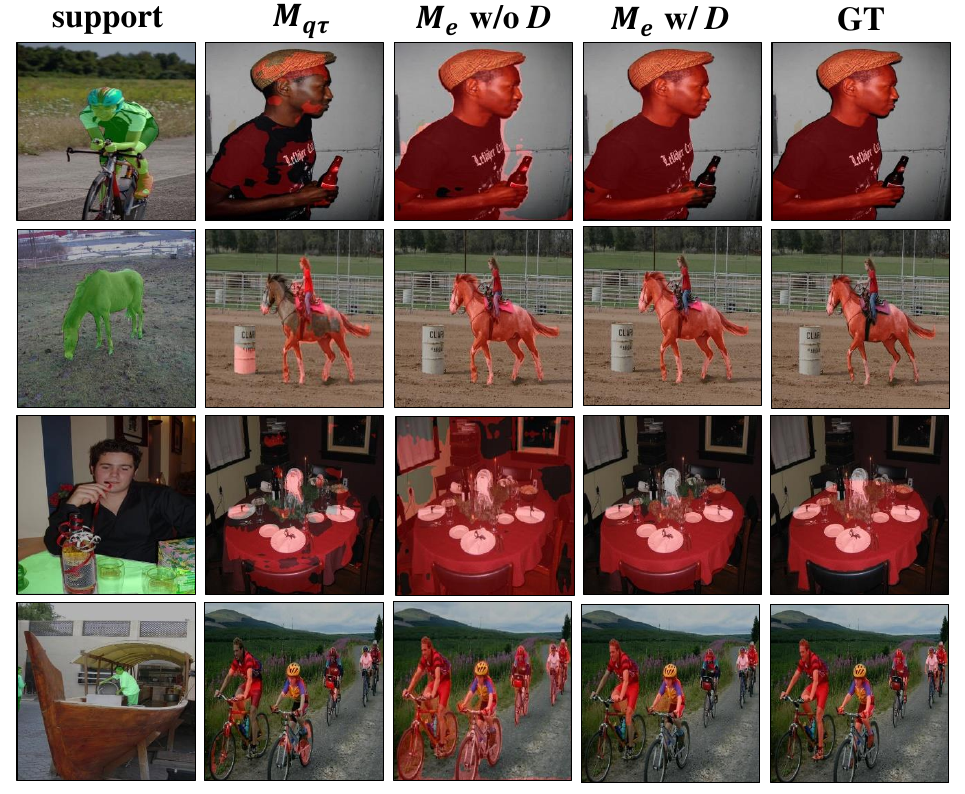}
    \caption{Qualitative results from different stages.}
    \label{fig:visual1}
    \end{minipage}
    %
    \hfill
    \begin{minipage}[t]{0.48\linewidth}
    \centering
    \includegraphics[width=1\linewidth]{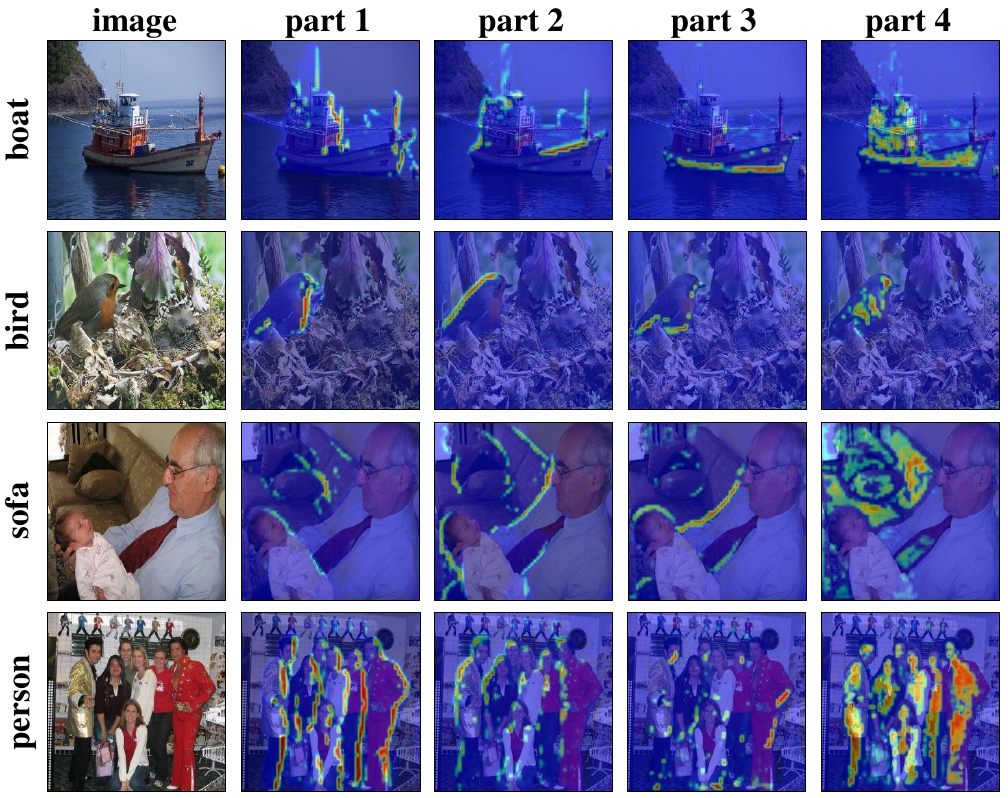}
    \caption{Visualization of regions activated by different local proxies.}
    \label{fig:visual3}
    \end{minipage}
    \vspace{-5mm}
\end{figure}

\begin{wrapfigure}{r}{0.38\textwidth}
    \centering
    \vspace{-8mm}
    \includegraphics[width=1.0\linewidth]{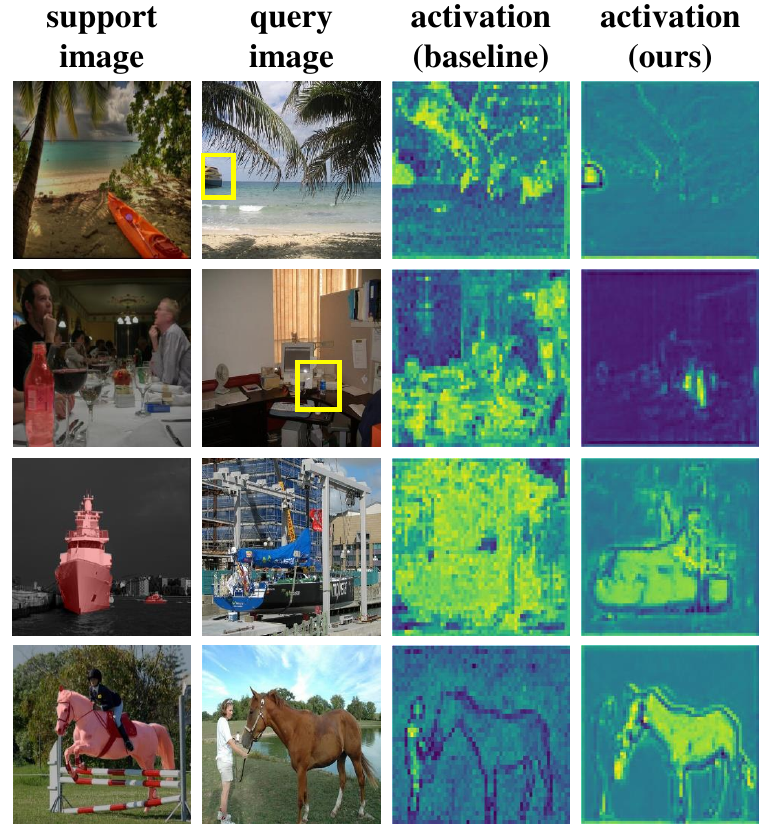}
    \vspace{-4mm}
    \caption{Visualization of the activation of query images..}
    \label{fig:visual2}
    \vspace{-3mm}
\end{wrapfigure}

\textbf{Investigation of the local proxies.}
We first visualize the region activated by the learnable proxies to analyze what they mainly focus on. As shown in Figure~\ref{fig:visual3}, we observe that the highlighted area mainly lies on the boundary of the target and different proxies correspond to different directions.
It confirms that the local proxies can well capture the local misalignment of the relatively coarse prediction from \textit{G}, and then \textit{G} is optimized to pay more attention to these ambiguous areas under the direction of \textit{D}.
In addition, we investigate the impact of the number of proxies on performance. As reported in Table~\ref{tab:part}, using more proxies can gradually improve the performance until the number achieves 10. More proxies no longer bring additional improvement. This result is expected, as too many proxies would learn redundant patterns.


\textbf{Discussion on the query-centric FSS paradigm.}
Since the FSS models are trained in the extremely low data regime with the backbone not optimized, the feature space is not well aligned for novel classes. In this situation, the novel class data distribution usually has low coverage in the feature space, \ie, object features of the same category but different image instances may be far apart.
On the contrary, as we can observe in Table~\ref{tab:inter-intra}, pixel features within the same object are closer to each other than those across objects. In a nutshell, intra-object similarity is more reliable than intra-category but inter-object similarity.
%
As shown in Figure~\ref{fig:visual2}, we visualize the attention weight of query features between support target and pseudo support, respectively. It shows that support targets tend to mistakenly activate background categories, while the pseudo support can well excavate the full object attribute to intra-object similarity.

Owing to the reduced reliance on support information, query-centric FSS methods could achieve remarkable performance with more weak support annotations. Table~\ref{tab:support} shows the results of the proposed AMFormer on the first split of Pascal-$5^i$ with bounding boxes or scribbles as support labels. The promising results demonstrate the feasibility of more general FSS segmentation models. 

\subsection{Broader Impact and Limitations.}
We proposed a novel query-centric FSS paradigm that shifts the research focus from support to query features. This is a new perspective that may inspire the development of more general FSS models that can be adopted in different tasks such as video object segmentation~\cite{cheng2021rethinking,sun2023alignment,sun20221st} or open-vocabulary segmentation~\cite{zhao2017open,xu2023side}. Although our AMFormer achieves remarkable performance, the number of training epochs is larger than some of previous approaches since the \textit{G} and \textit{D} in the AMFormer are trained alternately.


\vspace{-2mm}
\section{Conclusion}
\vspace{-2mm}
In this paper, we propose a novel query-centric FSS method, \ie, Adverasial Mining Transformer (AMFormer), which can achieve accurate query segmentation with only rough support guidance. Extensive experimental results demonstrate the superiority of our method. The decent performance with weak support labels also demonstrates the potential of the query-centric FSS paradigm.

\section{Acknowledgments}
This work was partially supported by the National Defense Basic Scientific Research Program of China (Grant JCKY2021601B013).


\bibliographystyle{unsrt}
\bibliography{egbib}

\newpage
\section{More Details for Multi-scale Object Mining Transformer.}
In the object mining transformer \textit{G}, we alternately use down-sampling layers and self-attention layers to construct hierarchical query features.
And the corresponding pseudo support features are obtained by the Hadamard product of the downsampled $M_{q\tau}$, specifically,
\begin{equation}
    \mathbf{F}_{q,l} = \mathbf{Down}(\mathcal{F}^{-1}(\mathbf{FeatAgg}(\mathcal{F}(\mathbf{F}_{q,l-1}),\mathcal{F}(\mathbf{F}_{q,l-1})))),
\end{equation}
where  the $\mathcal{F}: \mathbb{R}^{H \times W\times C} \mapsto \mathbb{R}^{HW\times C}$ is the spatial flatten operation and $\mathbf{Down}$ denotes the down-sampling layers, which is implemented with convolutional layers of double strides. In this way, we obtain multi-scale query features $\mathbf{F}_{q,l} \in \mathbb{R}^{\frac{H}{2^{l-1}} \times \frac{W}{2^{l-1}} \times C}, l=1,\dots,L$ and pseudo support features $\mathbf{F}_{psd,l} = \mathbf{F}_{q,l} \odot \varphi(M_{q\tau})$. ($\varphi$ is bilinear interpolation operation).
We perform feature aggregation within each scale to enable contextual information exploration, thus avoiding the spatial inconsistency in the query image.
$L$ is set to 3 in our experiments. The embedding dim is set to 64, and the number of head is set to 4 in all the attention layers.

\section{Detailed Experimental Settings}
To achieve a fair comparison with previous methods~\cite{lang2022learning, hdmnet}, we adopt the ensemble strategy following BAM~\cite{lang2022learning} to filter the base categories seen during training.
Specifically, a base learner is trained using the training splits in a fully supervised manner, and the learned base learner is used to explicitly predict the targets of base classes.
PSPNet~\cite{pspnet} is adopted as the base learner in all of our experiments, and we follow the BAM~\cite{lang2022learning} to ensemble the prediction of the base learner and the proposed AMFormer.
Our code is available at \url{https://github.com/Wyxdm/AMNet}

We  also conduct an additional ablation experiment to evaluate the influence of the ensemble strategy as shown in Table~\ref{tab:no ensemble}. We can observe that the ensemble strategy can incrementally improve performance. It should be noted that our AMFormer can also surpass previous state-of-the-art methods (IPMT~\cite{liu2022intermediate}) without the ensemble strategy.
\begin{table}[!thp]
    \caption{Ablation of ensemble strategy on Pascal-$5^i$ with ResNet-101 backbone and 1-shot setting.}
    \label{tab:no ensemble}
    \centering
    \begin{tabular}{c|ccccc}
        \toprule
         &fold0 & fold1 & fold2 & fold3 & mean   \\
                 \hline
                 \hline
IPMT~\cite{liu2022intermediate} &\textbf{71.6} &73.5 &58.0 &61.2 &66.1 \\
         w/o ensemble & 69.7 & \textbf{75.2} &\textbf{69.5} &\textbf{62.9}  &\textbf{69.3}      \\
         w/ ensemble  & \textcolor{gray}{71.3} & \textcolor{gray}{76.7} &\textcolor{gray}{70.7} &\textcolor{gray}{63.9}  &\textcolor{gray}{70.7}\\
         \bottomrule
    \end{tabular}
\end{table}

\section{Dataset Settings}
We respectively divided Pascal-5$^i$ and COCO-20$^i$ into four splits following~\cite{pfenet} for cross-validation. In Table~\ref{tab:pascal_splits} and Table~\ref{tab:coco_splits}, we provide the detailed split settings.
\begin{table}[!thp]
    \caption{Detailed splits setting of PASCAL-5$^i$}
    \label{tab:pascal_splits}
    \centering
    \begin{tabular}{c|c}
        \toprule
         Fold & Test classes \\
         \hline
         \hline
         PASCAL-5$^0$ & aeroplane, bicycle, bird, boat, bottle  \\
         PASCAL-5$^1$ & bus, car, cat, chair, cow  \\
         PASCAL-5$^2$ & diningtable, dog, horse, motorbike, person  \\
         PASCAL-5$^3$ & potted plant, sheep, sofa, train, tv/monitor \\
         \bottomrule
    \end{tabular}
\end{table}
 
\begin{table}[t]
    \caption{Detailed splits  setting of COCO-20$^i$}
    \label{tab:coco_splits}
    \centering
    \begin{tabular}{c|c}
        \toprule
         Fold & Test classes \\
         \hline
         \hline
         COCO-20$^0$ & \makecell{Person, Airplane, Boat, Park meter, Dog, Elephant, Backpack, Suitcase, \\ Sports ball, Skateboard, W. glass, Spoon, Sandwich, Hot dog, Chair, \\ D. table, Mouse, Microwave, Fridge, Scissors}  \\ \midrule
         COCO-20$^1$ & \makecell{Bicycle, Bus, T.light, Bench, Horse, Bear, Umbrella, \\ Frisbee, Kite, Surfboard, Cup, Bowl, Orange, Pizza, \\ Couch, Toilet, Remote, Oven, Book, Teddy} \\ \midrule
         COCO-20$^2$ & \makecell{Car, Train, Fire H., Bird, Sheep, Zebra, Handbag, \\ Skis, B. bat, T. racket, Fork, Banana, Broccoli, Donut, \\ P. plant, TV, Keyboard, Toaster, Clock, Hairdrier} \\ \midrule
         COCO-20$^3$ & \makecell{Motorcycle, Truck, Stop, Cat, Cow, Giraffe, Tie, \\ Snowboard, B. glove, Bottle, Knife, Apple, Carrot, Cake, \\ Bed, Laptop, Cellphone, Sink, Vase, Toothbrush} \\
         \bottomrule
    \end{tabular}
    \vspace{-3mm}
\end{table}

\section{More Experimental Results}
\subsection{Quantitative analysis of intra-object similarity.} 
We compute the average pairwise pixel similarity from the same object (intra-object) and different objects from the support and query images of the same category (inter-object) using the cosine similarity. Note that the pixel features that we used to compute the similarity are middle-level features $\mathbf{F}_s$ and $\mathbf{F}_q$ as described in the L255-L256 in the original manuscript. The quantitative results of different categories are provided in Table~\ref{tab:intra_inter_pascal} and Table~\ref{tab:intra_inter_coco}.
From the tables, we can observe that the intra-object similarity is at least one order of magnitude higher than the inter-object similarity. This demonstrates the superiority of the query-centric approach relying on intra-object similarity over support-centric methods that rely on inter-object similarity.
\begin{table}[!thp]
    \caption{Intra- and inter-object similarity of each class within Pascal-5$^i$, }
    \label{tab:intra_inter_pascal}
    \centering
    \tabcolsep 0.04in\begin{tabular}{c|cc|c|cc|c|cc|c|cc}
			\bottomrule
			\multicolumn{3}{c|}{Pascal-5$^0$}  &  \multicolumn{3}{c|}{Pascal-5$^1$} &\multicolumn{3}{c|}{Pascal-5$^2$} & \multicolumn{3}{c}{Pascal-5$^3$}   \\
			\hline
			class & Intra-  &  Inter- & class & Intra-  &  Inter- &class & Intra-  &  Inter- &class & Intra-  &  Inter- \\
			\hline
			\hline
			 aeroplane &0.449 &0.008 &bus &0.453 &0.011 &diningtable &0.496 &0.010 &potted plant &0.515 &0.012 \\
			 bicycle &0.460 &0.009 &bus &0.453 &0.011 &dog &0.571 &0.009 &sheep &0.536 &0.007 \\
			 bird &0.493 &0.010 &cat &0.643 &0.021 &horse &0.489 &0.012 &sofa &0.535 &0.006 \\
			 boat &0.483 &0.009 &chair &0.519 &0.015 &motorbike &0.445 &0.008 &train &0.509 &0.008 \\
			 bottle &0.504 &0.120 &cow &0.573 &0.016 &person &0.529 &0.039 &tv/monitor &0.513 &0.024 \\
			\bottomrule
	\end{tabular}
 \vspace{-3mm}
\end{table}

\begin{table}[!thp]
    \caption{Intra- and inter-object similarity of each class within COCO-20$^i$}
    \label{tab:intra_inter_coco}
    \centering
    \tabcolsep 0.027in\begin{tabular}{c|cc|c|cc|c|cc|c|cc}
			\bottomrule
			\multicolumn{3}{c|}{COCO-20$^0$}  &  \multicolumn{3}{c|}{COCO-20$^1$} &\multicolumn{3}{c|}{COCO-20$^2$} & \multicolumn{3}{c}{COCO-20$^3$}   \\
			\hline
			class & Intra-  &  Inter- & class & Intra-  &  Inter- &class & Intra-  &  Inter- &class & Intra-  &  Inter- \\
			\hline
			\hline
			 Person&0.553 &0.002 & Bicycle &0.398 &0.017 &Car &0.514 &0.012 &Motorcycle &0.485 &0.007 \\
			 Airplane&0.582 &0.020 &Bus &0.440 &0.019 &Train &0.544 &0.017 &Truck &0.552 &0.021 \\
			 Boat&0.585 &0.013 &T.light &0.474 &0.013 &Fire H. &0.467 &0.007 &Stop &0.544 &0.008 \\
			 Park meter&0.476 &0.014 &Bench &0.459 &0.044 &Bird &0.530 &0.016 &Cat &0.549 &0.010 \\
			 Dog&0.482 &0.010 &Horse &0.574 &0.018 &Sheep &0.487 &0.010 &Cow &0.584 &0.015 \\
             Elephant&0.467 &0.012 &Bear &0.460 &0.012 &Zebra &0.510 &0.012 &Giraffe &0.556 &0.013 \\
             Backpack&0.479 &0.013 &Umbrella &0.571 &0.020 &Handbag &0.519 &0.032 &Tie &0.559 &0.006 \\
             Suitcase&0.542 &0.007 &Frisbee &0.384 &0.005 &Skis &0.432 &0.007 &Snowboard &0.509 &0.014 \\
             Sports ball&0.538 &0.009 &Kite &0.470 &0.016 &B.bat &0.532 &0.023 &B.glove &0.576 &0.006 \\
             Skateboard&0.537 &0.018 &Surfboard &0.589 &0.024 &T.racket &0.373 &0.012 &Bottle &0.545 &0.007 \\
             W.glass&0.434 &0.011 &Cup &0.606 &0.009 &Fork &0.451 &0.017 &Knife &0.589 &0.013 \\
             Spoon&0.562 &0.027 &Bowl &0.563 &0.010 &Banana &0.511 &0.015 &Apple &0.656 &0.010 \\
             Sandwich&0.469 &0.013 &Orange &0.474 &0.028 &Broccoli &0.471 &0.026 &Carrot &0.539 &0.007 \\
             Hot dog&0.558 &0.008 &Pizza &0.515 &0.015 &Donut &0.477 &0.300 &Cake &0.597 &0.006 \\
             Chair&0.442 &0.015 &Couch &0.542 &0.013 &P.plant &0.433 &0.032 &Bed &0.560 &0.007 \\
             D.table&0.474 &0.046 &Toilet &0.503 &0.006 &TV &0.456 &0.014 &Laptop &0.480 &0.021 \\
             Mouse&0.447 &0.021 &Remote &0.583 &0.011 &Keyboard &0.542 &0.007 &Cellphone &0.498 &0.018 \\
             Microwave&0.454 &0.019 &Oven &0.567 &0.011 &Toaster &0.433 &0.007 &Sink &0.375 &0.027 \\
             Fridge&0.476 &0.023 &Book &0.586 &0.025 &Clock &0.526 &0.018 &Vase &0.567 &0.013 \\
             Scissors&0.456 &0.027 &Teddy &0.583 &0.016 &Hairdrier &0.373 &0.007 &Toothbrush &0.479 &0.013 \\
			\bottomrule
	\end{tabular}
 \vspace{-3mm}
\end{table}

\subsection{More visualization results.}
\begin{figure}[t!]
    \centering
    \includegraphics[width=\linewidth]{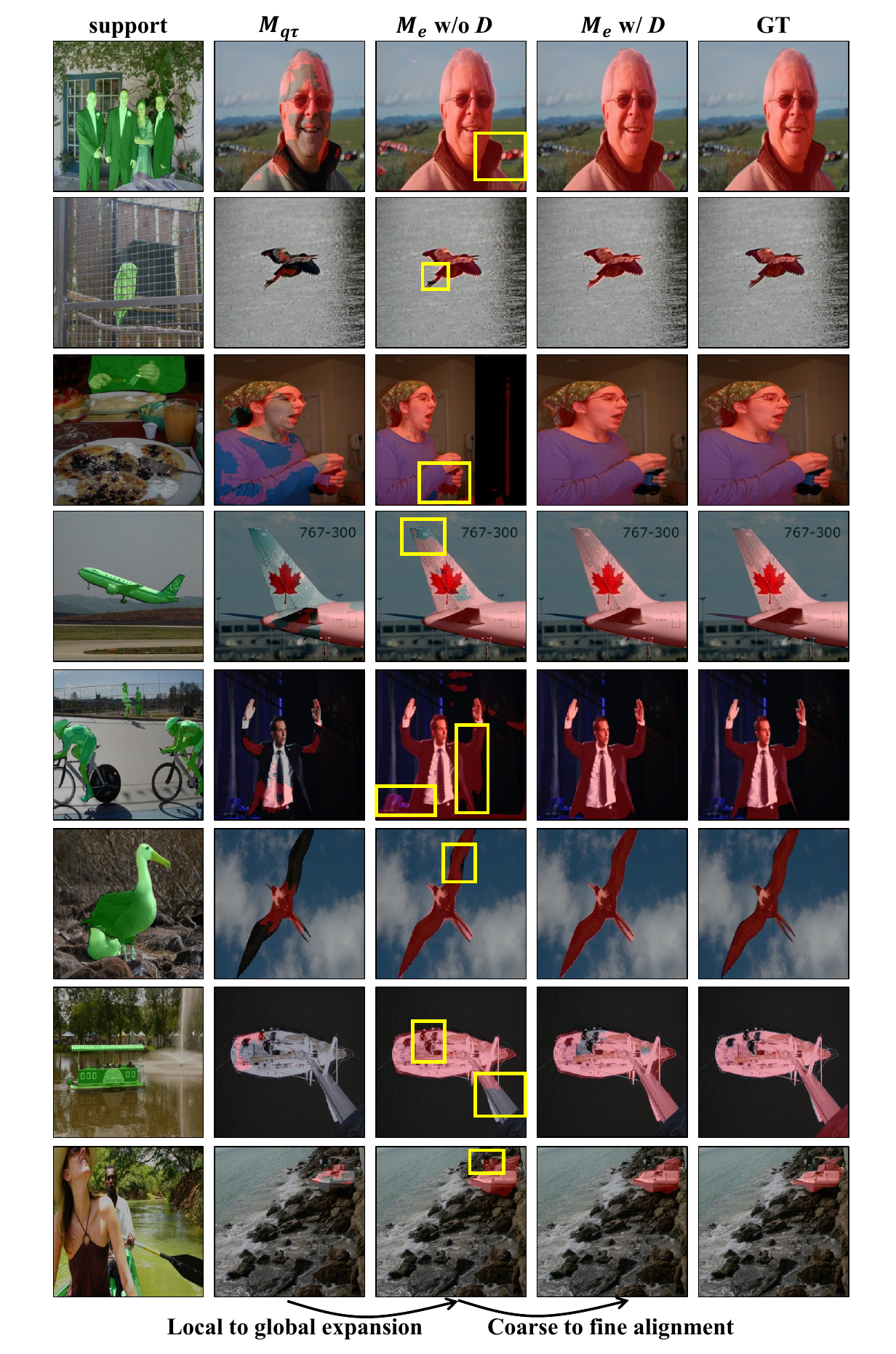}
    \caption{Visualization of the segmentation of the proposed AMFormer at different stages}
    \label{fig:final}
    \vspace{-5pt}
\end{figure}

\textbf{Visualization of segmentation at different stages.}
We tackle query-centric FSS by applying three intuitive procedures. (1) Discriminative region localization. (2) Local to global expansion. (3) Coarse to fine alignment. To illustrate the effects of the above three steps, in Figure~\ref{fig:final}, we visualize the outcomes of different stages. Procedure (1) can only roughly local the discriminative region of the target ($2^{nd}$ column of Figure~\ref{fig:final}). In procedure (2), the object mining transformer \textit{G} exploits the intra-object similarity to explore multi-scale contextual information, thus highlighting the whole object ($3^{rd}$ column of Figure~\ref{fig:final}).
Segmentation from \textit{G} can roughly cover the entire target but there still exist misalignments as shown in the yellow boxes in the $3^{rd}$ column of Figure~\ref{fig:final}.
The detail mining transformer \textit{D} is responsible for discriminating those detailed misalignments, \ie, procedure (3).
The proposed AMFormer couples procedures (2) and (3) via adversarial training. In this way, the G can be optimized to generate more accurate segmentations($4^{th}$ column of Figure~\ref{fig:final}) to fool \textit{D}.

\begin{figure}[t!]
    \centering
    \includegraphics[width=\linewidth]{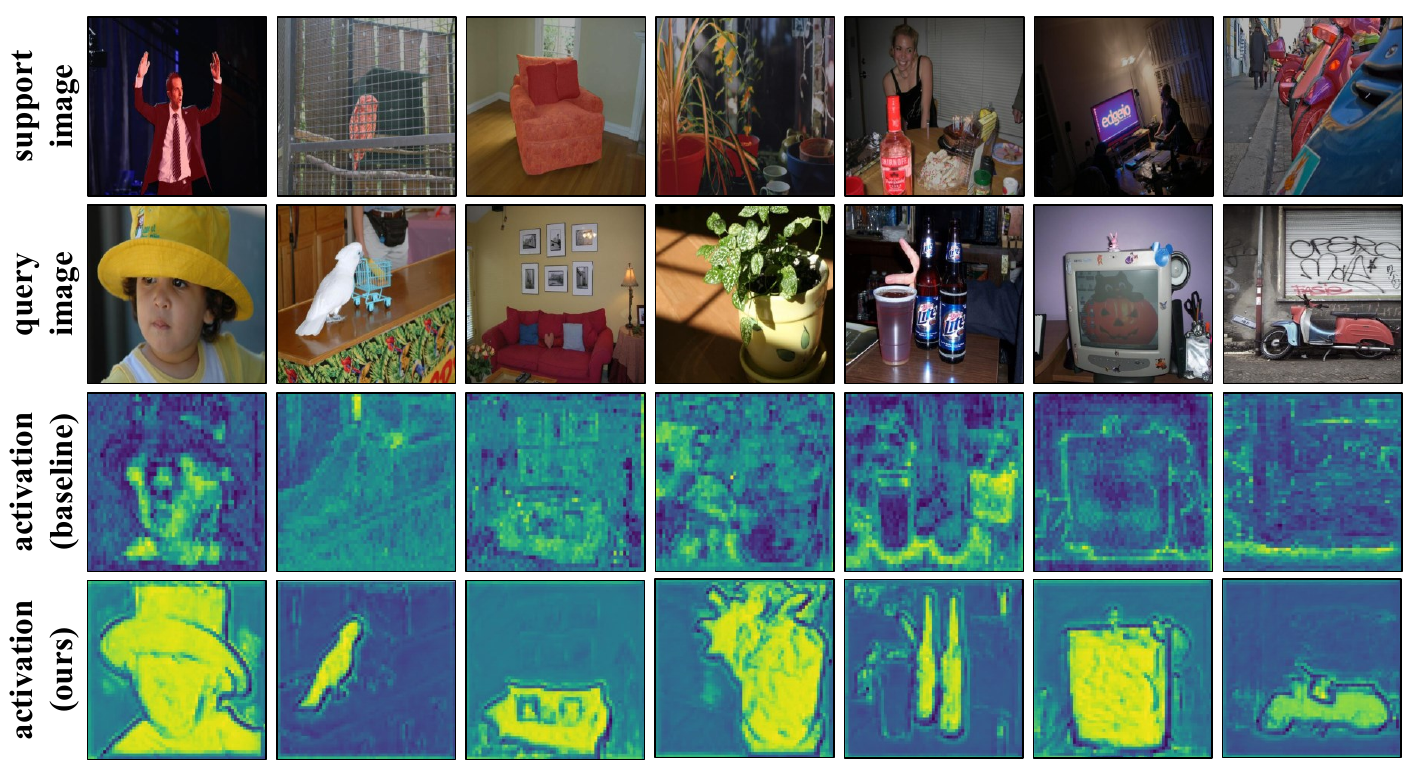}
    \caption{Visualizations of the attention weight of query features between the support target (support-centric baseline) and pseudo support (ours).}
    \label{fig:activation}
    \vspace{-5pt}
\end{figure}

\textbf{Visualization of activation maps.} We visualize the attention weight of query features between the support target and pseudo support. Specifically, the attention matrix $\mathbf{S} \in \mathbb{R}^{H_qW_q \times H_sW_s}$ is computed according to the Eqn (1) of the original manuscript. Then we compute the average activation of each query pixel over all support foreground pixels:
\begin{equation}
    \mathbf{Act}(i) = \frac{\sum_{j=1}^{H_sW_s}\mathbf{S}(i,j)\cdot[\mathcal{F}(\mathbf{M}_s)(j)>0]}{\sum_{j=1}^{H_sW_s}[\mathcal{F}(\mathbf{M}_s)(j)>0]},
    \label{eq:activation}
\end{equation}
where the $\mathbf{M}_s$ is the (pseudo) support mask. In the baseline, the $\mathbf{S}$ is oriented from the cross attention between the query features and the support features (support-centric). While in our query-centric AMFormer, the $\mathbf{S}$ is computed from the pseudo support and the query features.
As shown in Figure~\ref{fig:activation}, the support targets not only cannot fully activate the target in the query image, but also frequently activates the background categories.
While the pseudo support can well excavate the full object attribute to intra-object similarity.

\begin{figure}[t!]
    \centering
    \includegraphics[width=\linewidth]{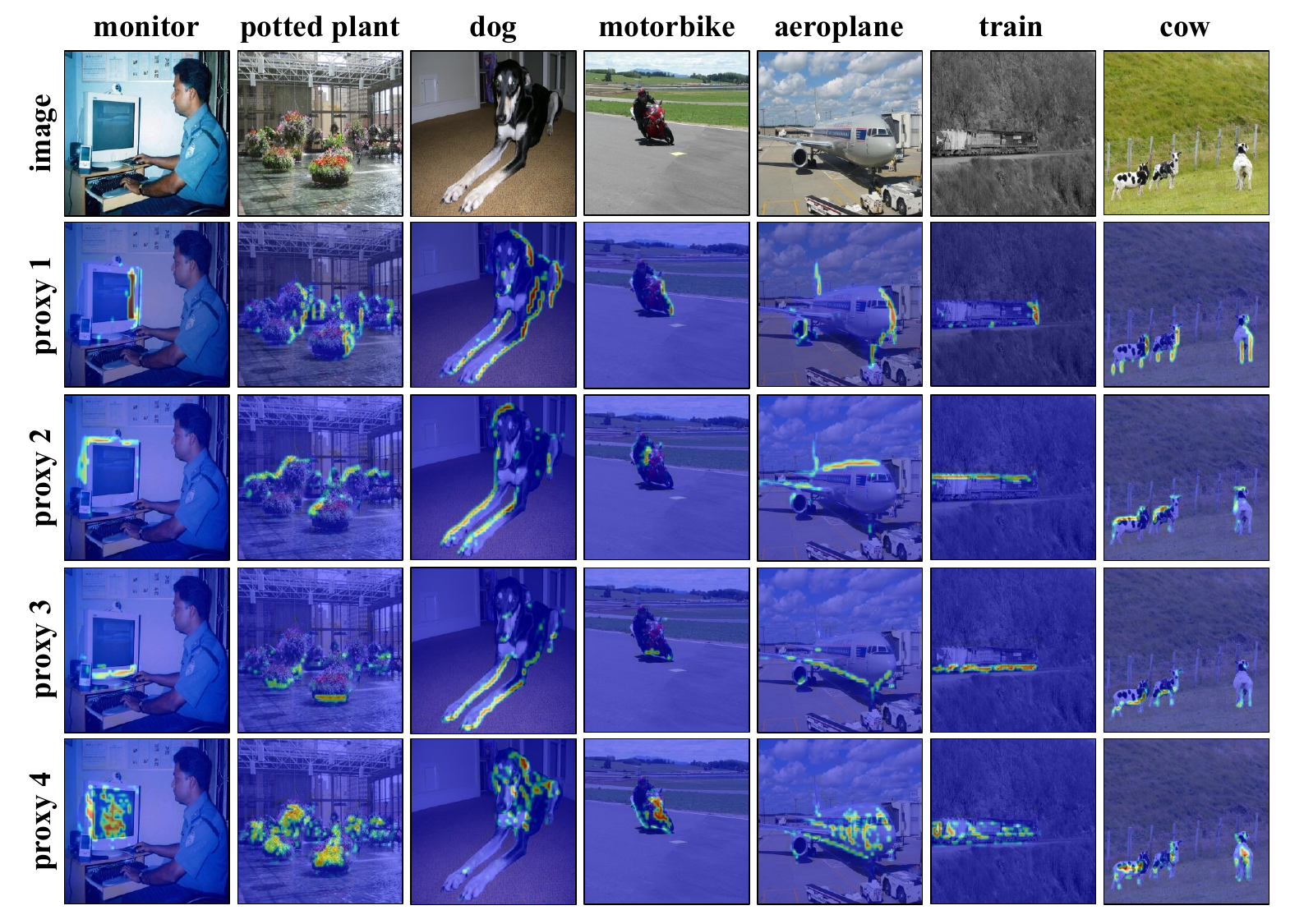}
    \caption{Visualizations of the activated regions of local proxies.}
    \label{fig:part}
    \vspace{-5pt}
\end{figure}

\textbf{Visualization of local proxies.} To explore the regions of interest for learnable local proxies, Figure~\ref{fig:part} visualizes the activation maps of a part of proxies. It can be observed that different proxies tend to focus on different local regions, and most proxies attend to the boundaries of the object, which is usually the most ambiguous region.
In addition, a particular proxy consistently focuses on the boundaries in a specific direction, \eg, \textit{proxy 1} always activates the right border~($5^{th} $ column).
Through the cooperation of multiple proxies, our detail mining transformer \textit{G} can effectively detect the detailed local differences between the prediction of the object mining transformer \textit{G} and the ground truth.
By means of adversarial training, G will produce more accurate segmentations, especially in ambiguous regions, by adjusting itself to fool D.

\end{document}